\documentclass{article}

\usepackage{PRIMEarxiv}

\usepackage[utf8]{inputenc} 
\usepackage[T1]{fontenc}    
\usepackage{hyperref}       
\usepackage{url}            
\usepackage{booktabs}       
\usepackage{amsfonts}       
\usepackage{nicefrac}       
\usepackage{microtype}      
\usepackage{lipsum}
\usepackage{fancyhdr}       
\usepackage{graphicx}       
\usepackage{tikz}
\usepackage{marvosym}
\usetikzlibrary{trees, positioning}
\graphicspath{{media/}}     

\title{Investigating Public Fine-Tuning Datasets: A Complex Review of Current Practices from a Construction Perspective

}

\author{
  Runyuan Ma\thanks{First author.} \\
  Shanghai AI Laboratory \\
  Shanghai\\
  \texttt{marunyuan@pjlab.org.cn} \\
   \And
  Wei Li\thanks{Corresponding author.} \\
  Shanghai AI Laboratory \\
  Shanghai\\
  \texttt{liwei@pjlab.org.cn} \\
  \And
  Fukai Shang\\
  Shanghai AI Laboratory \\
  Shanghai\\
  \texttt{shangfukai@pjlab.org.cn} \\
}

\begin{document}
\maketitle

\begin{abstract}
With the rapid development of the large model domain, research related to fine-tuning has concurrently seen significant advancement, given that fine-tuning is a constituent part of the training process for large-scale models. Data engineering plays a fundamental role in the training process of models, which includes data infrastructure, data processing, etc. Data during fine-tuning likewise forms the base for large models. In order to embrace the power and explore new possibilities of fine-tuning datasets, this paper reviews current public fine-tuning datasets from the perspective of data construction. An overview of public fine-tuning datasets from two sides: evolution and taxonomy, is provided in this review, aiming to chart the development trajectory. Construction techniques and methods for public fine-tuning datasets of Large Language Models (LLMs), including data generation and data augmentation among others, are detailed. This elaboration follows the aforementioned taxonomy, specifically across demonstration, comparison, and generalist categories. Additionally, a category tree of data generation techniques has been abstracted in our review to assist researchers in gaining a deeper understanding of fine-tuning datasets from the construction dimension. Our review also summarizes the construction features in different data preparation phases of current practices in this field, aiming to provide a comprehensive overview and inform future research. Fine-tuning dataset practices, encompassing various data modalities, are also discussed from a construction perspective in our review. Towards the end of the article, we offer insights and considerations regarding the future construction and developments of fine-tuning datasets.
\end{abstract}


\section{Introduction}
The field of large model research has experienced a year of dynamic engagement and substantial progress. Numerous open-source and closed-source foundation models, as well as fine-tuning (FT) models, have been developed globally. In the technological pathway from foundation models to fine-tuning models, high-quality and effective fine-tuning datasets are the foundation. Some fine-tuning models based on foundation models have made their fine-tuning datasets publicly available. Additionally, many large model projects have disclosed public datasets, a portion of which are dedicated to fine-tuning.  In the rapidly developing field of large model fine-tuning, the release of fine-tuning datasets not only enhances the influence of models and projects but also promotes fine-tuning research, facilitating the democratization of technological advancements.\\
Currently, there is a lack of dedicated and systematic reviews concerning the public datasets applied to the fine-tuning stage. Existing discussions on fine-tuning datasets often appear as supplements to the release of dataset generalists or the data collection part of specific models, leaving a gap in specialized reviews focused on large model fine-tuning datasets. Some studies in the field of public fine-tuning datasets tend to primarily focus on the statistical aspects of these datasets while lacking a comprehensive analysis of the longitudinal evolution and key milestones of fine-tuning dataset development. Meanwhile, certain research works introducing fine-tuning datasets are no longer at the cutting edge, while advancements in the field of public fine-tuning datasets are progressing on a quick cadence. Hence, a comprehensive and up-to-date review of fine-tuning dataset trends and developments is urgently needed to provide an overview.\\
Fine-tuning data is defined as the data applied during the fine-tuning phase of large models, the scope of which depends on the delineation of the fine-tuning. We draw inspiration from the fine-tuning steps outlined in OpenAI's InstructGPT paper\cite{ouyang2022training}, encompassing Supervised Fine-Tuning (SFT)and Preference Modeling (RM, PPO), rather than solely focusing on the SFT step. In this review, according to the scope defined for fine-tuning, the range of fine-tuning data includes not only the data used to leverage the large model's instruction-following capabilities but also focuses on all data used to finetune language models to align large models with human behaviors, such as demonstration data and comparison data.\\
This paper reviews the fine-tuning datasets, focusing on understanding the development map of these datasets from a construction perspective and prioritizing attention to publicly available fine-tuning datasets.   We have adopted a unique taxonomy for fine-tuning datasets within the defined scope of Large Language Models (LLMs), reviewed the construction methods of various LLM fine-tuning datasets, and discussed the commonalities and differences in construction methods across different categories of datasets, thereby substantiating the rationale behind their evolution and taxonomy. Attention is then shifted to fine-tuning datasets in the domain of multimodal large models for further investigation. In conclusion, future directions for the development of large-scale model fine-tuning datasets are proposed, aiming to provide a higher quality fine-tuning dataset paradigm for large model learning.
\paragraph{Contribution}
\begin{itemize}
\item The democratization process of fine-tuning dataset research with datasets as the primary entity
\item A comprehensive analysis of public fine-tuning datasets: encompassing supervised fine-tuning (SFT) and preference modeling phases across textual and multimodal modalities
\item Construction techniques, methods, and a categorization framework for data generation techniques in the field of public fine-tuning datasets
\end{itemize}

\section{Overview}
\label{sec:Overview}
\subsection{Evolution}
The evolution of fine-tuning datasets can be categorized into distinct phases over two generations. Notably, InstructGPT\cite{ouyang2022training} stands as a landmark achievement in the realm of large models and, more specifically, fine-tuning data. We regard the release of the InstructGPT\cite{ouyang2022training} paper as a watershed moment in the domain of fine-tuning research. Prior to this pivotal development, the fine-tuning dataset landscape was significantly distinct.

\paragraph{The previous generation: Aggregation of NLP Tasks}
When it comes to data for large model fine-tuning phases, fine-tuning datasets have evolved from the foundation of Natural Language Processing(NLP) data. In the previous generation, the primary methodology for creating fine-tuning datasets involves aggregating tasks in the field of NLP and converting them into instruction formats. Subsequently, these fine-tuning datasets, obtained in this manner, are combined with tasks into a single data source. The landscape of NLP task aggregation has been shaped by existing studies, with our thoughts being particularly inspired by the contributions of The Flan Collection\cite{longpre2023flan} (2023).\\
The review and analysis of fine-tuning datasets during this generation received relatively less attention in this article.
\paragraph{The later generation: Arrival of InstructGPT\cite{ouyang2022training}}
InstructGPT\cite{ouyang2022training} is a notable project in this domain. InstructGPT\cite{ouyang2022training} and the stages that followed focus on doing general tasks using instructions that are more natural and interactive. Furthermore, it introduces a novel perspective on the fine-tuning stage of LLMs, as well as the data collection process used in each technique. However, the creation of data at each stage in OpenAI's InstructGPT\cite{ouyang2022training} technical report involves a lot of human labor, including user contributions in the so-called automatic API collections.
In the following days, the Emergence of Synthesis-Paradigm Datasets gives new momentum to the field of fine-tuning datasets. 
Datasets divided into this generation receive significant attention in this review.
\subsection{Taxonomy} 
A set of criteria has been defined to subjectively categorize and understand fine-tuning datasets. We use modality, and purpose as orthogonal criteria. Under each criterion, different categories will be identified.

\paragraph{Modality.}The modality of the fine-tuning datasets is regarded as the most basic classification criterion in this review. Modality itself always serves as an essential and significant attribute of the data. For the presently existing large-scale language models, fine-tuning datasets primarily exist in an unimodal form, specifically as a single-text modality. The emergence of Multimodal Large Language Models (MLLMs) has ushered in the appearance of fine-tuning datasets with multiple modalities. Currently, in the democratization process of fine-tuning dataset research, NLP fine-tuning datasets for large language models dominate, while fine-tuning datasets dedicated to MLLMs are emerging. Hence we raise modality as the Primary Classification Criteria.
\paragraph{Purpose.}Datasets are categorized as "Demonstration Datasets", "Comparison Datasets", and "Generalist Datasets" to reflect their diverse purposes in research. The nomenclature "demonstration" and "comparison" follow the terminology established in the InstructGPT\cite{ouyang2022training} paper. While most datasets predominantly consist of either demonstration or comparison data, with the advancement of research in the fine-tuning dataset field, some datasets now incorporate both demonstration and comparison data or individual data points within a dataset serve dual roles, supporting both the Supervised Fine-Tuning and Preference Modeling stages. We also define datasets of this nature as "Comparison Datasets." The term "Generalist" is adopted from the naming convention introduced in the project"Open Instruction Generalist"\cite{oig2023}. \\Pre-trained models, after acquiring a wealth of world knowledge during the pre-training phase, face the challenge of aligning themselves with human cognition. This alignment occurs along two dimensions: aligning with the way humans communicate in language and aligning with human value preferences. In the process of building large-scale models, some alignment work is conducted during fine-tuning. The selection and construction of appropriate fine-tuning datasets play a pivotal role in injecting and amplifying human intelligence into the model. Demonstration data informs the model about patterns for answering specific questions, while comparison data guides the model toward thinking more closely in line with human reasoning. Fine-tuning datasets encompass at least one of these two types of data. Next, a concise overview of each of these dataset categories will be provided.

\section{Fine-tuning Dataset For LLMs}
\label{sec:others}

The methodology employed in constructing datasets contributes a lot to their quality. Datasets can be constructed through various methods, and these different construction approaches vary in terms of data sources, data collection, data annotation, construction costs, and so on. Fine-tuning datasets are also affected by their construction methods.  Fine-tuning datasets applied to pre-trained models have influenced the performance of large models. Compared to the pretraining data of large models, fine-tuning data places finer demands on annotation and undergoes rigorous quality control. Understanding the diverse construction methods of fine-tuning datasets can lead to a better comprehension of these datasets and support more effective fine-tuning dataset design, thereby facilitating advancements in the field of fine-tuning dataset research.\\
The methods for constructing high-quality fine-tuning datasets vary case by case. However, some common key techniques can be extracted from existing open-source fine-tuning datasets. In the paragraphs below, our paper reviews key techniques of data-building methods. \\
The construction of fine-tuning datasets is often a multi-module process, where each construction method is not isolated, and each fine-tuning dataset may involve multiple data construction methods. There are often various approaches along the road to the terminal point of dataset construction. Some popular pipelines integrating multiple steps are also highlighted in the remainder of this paper. Additionally, the various construction methods for fine-tuning datasets did not emerge simultaneously; they have distinct origins and developmental trajectories. To gain a deeper understanding of current fine-tuning datasets from a construction perspective, it is essential to trace the evolutionary history of different methods and analyze their development over time. In this context, we make efforts to dissect the development timeline of fine-tuning dataset construction methods, examine the mainstream construction strategies and their characteristics at different stages, and explore the connections and evolution among various methods.\\

Adopted from the Section\ref{sec:Overview}, demonstration datasets, comparison datasets and categories left will be described separately next.
\subsection{Construction of Demonstration Datasets}
Based on the methodology of data construction, the preparation of demonstration data can be categorized into two primary approaches: data generation and data augmentation. Data generation itself encompasses two subtypes: human-generated data and synthesized data, with the latter predominantly relying on model-generated techniques. Data augmentation, on the other hand, primarily involves the reformatting of pre-existing public datasets. It is pertinent to note that if an instruction from a public dataset results in new data via data generation techniques, this is classified as data generation. In contrast, when both the instruction and response are from a public dataset and involve data augmentation, the scenario is categorized as data augmentation. To provide a clear overview of the construction features of public fine-tuning demonstration datasets, Table 1 \ref{tab: summary table}presents a summary of their key methods and techniques. The detailed descriptions of these methods and techniques will be provided in the following paragraphs.\\

\begin{table}
 \caption{Summary table of public fine-tuning demonstration datasets: Construction methods and techniques.}
  \centering
  \begin{tabular}{p{2.4cm}p{1.3cm}p{0.9cm}p{3.5cm}p{3.3cm}p{1.8cm}}
    \toprule
    \multicolumn{1}{l}{Name} & \multicolumn{5}{l}{Construction}  \\
    \cmidrule(r){2-6}
    & Method & Group & Inst. Preparation & Resp. generation & Note \\
    \midrule
    Unnatural Instructions\cite{honovich-etal-2023-unnatural} & Generated & M-M&Model-Generated/davinci & Model-Generated/davinci &  \\
    Self-Instruct\cite{wang-etal-2023-self-instruct}   & Generated &M-M &Model-Generated/text-davinci-003 &Model-Generated/text-davinci-003  &   \\
    Alpaca\cite{taori2023alpaca} & Generated &M-M & &  &   \\
    Alpaca\_gpt4\_data\cite{Peng2023InstructionTW} & Generated & M-M&Model-Generated/davinci & Model-Generated/davinci &  \\  
    InstructWildv1\cite{instructionwild}& Generated &M-M&Model-Generated/davinci & Model-Generated/davinci &  \\
    InstructWildv2\cite{instructionwild} & Generated & H-H&Human-Generated/user-based/user shared & - &  \\
    ShareGPT90K\cite{ShareGPT}&Generated&H-M&Human-Generated/user-based/user shared&Model-Generated/ChatGPT&-\\
    belle/data/1.5M\cite{BELLE,ji2023exploringimpactinstructiondata,wen2023chathome}&Generated&M-M&Model-Generated/ChatGPT&Model-Generated/ChatGPT&Self-Instruct Style \\
    Databricks-dolly-15k\cite{databricks2023}&Generated&H-H&Human-Generated/Crowdsourcing&Human-Generated/Crowdsourcing&Inspired by InstructGPT\\
    LIMA\cite{NEURIPS2023_ac662d74}&Generated&H-H&Human-Generated&Human-Generated&-\\
    OL-CC\cite{ol-cc2023}&Generated&H-H&Human-Generated/Crowdsourcing&Human-Generated/Crowdsourcing&-\\
    No Robots\cite{no_robots}&Generated&H-H&Human-Generated/Crowdsourcing&Human-Generated/Crowdsourcing&Inspired by InstructGPT\\
    UltraChat\cite{ding-etal-2023-enhancing}&Generated&M-M&Model-Generated/ChatGPT Turbo API&Model-Generated/ChatGPT Turbo API&-\\
    ChatAlpaca\cite{ChatAlpaca}&Generated&M-M&Model-Generated/using Alpaca data&Model-Generated/GPT-3.5-turbo&\\
    Code Alpaca\cite{codealpaca}&Generated&M-M&Model-Generated/text-davinci-003&Model-Generated/text-davinci-003&modified Stanford Alpaca\\
    Baize\cite{xu2023baize}&Generated&M-M&&\\
    moss-002-sft-data\cite{sun2023moss}&Generated&H-M&From Anthropic red teaming data &model-generated/text-davinci-003&Self-Chat Style\\
    moss-003-sft-data\cite{sun2023moss}&Generated&M-M,H-M&Model-Generated/gpt-3.5-turbo,Human-Generated/user-based/ API&model-generated/MOSS& \\
    moss-003-sft-plugin-data\cite{sun2023moss}&Generated&H-M&Human-Generated/user-based/API&model-generated/MOSS&\\
    firefly-train-1.1M\cite{Firefly}&Augmented& & & &  \\
    Open-Platypus\cite{lee2024platypus}&Augmented&- &-&-& -\\
    MathInstruct\cite{yue2023mammothbuildingmathgeneralist}&Augmented& -&-&-&- \\
    MetaMathQA\cite{yu2023metamath}&Augmented&- &-&-&-\\
    \bottomrule
  \end{tabular}
  \label{tab: summary table}
\end{table}

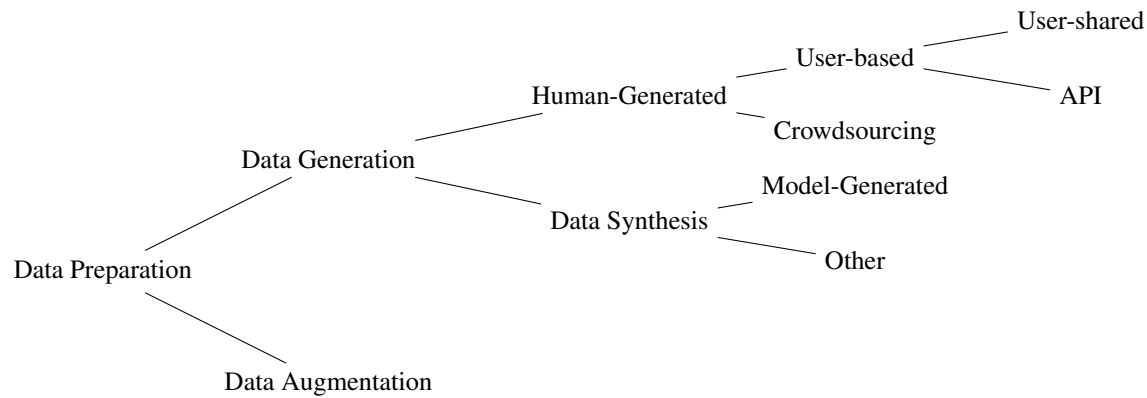
\begin{figure}
  \centering
  \begin{tikzpicture}[
    grow=right, 
    level 1/.style={sibling distance=3cm, level distance=3cm},
    level 2/.style={sibling distance=1.7cm, level distance=4cm},
    level 3/.style={sibling distance=1cm, level distance=3cm}
  ]

  \node {Data Preparation}
    child {
      node {Data Augmentation}
    }
    child {
      node {Data Generation}
      child {
        node {Data Synthesis}
        child {
          node {Other}
        }
        child {
          node {Model-Generated}
        }
      }
      child {
        node {Human-Generated}
        child {
          node {Crowdsourcing}
        }
        child {
          node {User-based}
          child {
            node {API}
          }
          child {
            node {User-shared}
          }
        }
      }
    };

  \end{tikzpicture}
  \caption{Key techniques Of Data Preparation.}
  \label{fig:Key techniques}
\end{figure}

In constructing fine-tuning data for large language models, the focus shifts significantly towards generating original data. This marks a shift from traditional AI data collection methods. Additionally, the data synthesis approach for fine-tuning public datasets is primarily model-based. The various techniques of data generation can be categorized into human-generated and model-generated methods. The method of generating demonstration datasets has changed over time, starting with human labor and later incorporating model distillation, going back and forth between human-generated and model-generated approaches. The following paragraphs will include dominant methods such as crowdsourcing, API use, user-sharing, and the Self-Instruct\cite{wang-etal-2023-self-instruct} style, among others. A clear visual representation of key construction techniques is showcased within a tree diagram\ref{fig:Key techniques}.

\paragraph{Human-Generated Technique}
\subparagraph{User-based}
API: API data collection is a popular practice for collecting user-submitted instructions after model deployment, including InstructGPT\cite{ouyang2022training}, ChatGPT, and GPT4. When collecting relevant instructions using an API, users must be informed that their submitted data will be used to update the model and that they must agree to the relevant agreement before using this data. Additionally, to avoid the model learning user privacy, instructions containing personal identity information must be filtered out. Furthermore, some ambiguous instructions need to be filtered manually, and related instructions need to be classified, such as helpfulness-related instructions or harmfulness-related instructions.\\
For API-collected instruction prompts, as part of the instruction prompts, the next step is for annotators to write responses that meet human desires to form the final instruction-response pair dataset. It can be said that collecting user instructions through APIs and rewriting responses through crowdsourcing are two complementary key links in the construction of instruction-response pair datasets. API collection emphasizes covering a wide range of instruction corpora, while manual response writing ensures the applicability and compliance of responses with human interaction needs.\\

User-shared: Users manually write prompts and share both the prompts and model responses through LLMs usage. One approach is collecting user-shared prompts and model-generated responses from public general online platforms. For example, InstructWild v2 in InstructionWild\cite{instructionwild} comprises over 110,000 used-based instructions from Twitter, Cookup.AI, GitHub, and Discord. Another approach is directly scraping prompt-response pairs from specialized websites where users post and share their conversations, such as the ShareGPT90k dataset collected from the ShareGPT\cite{ShareGPT} website, where users upload interesting GPT dialogues. Open-source large language models like Vicuna\cite{vicuna2023} used transcripts of ChatGPT conversations from the ShareGPT\cite{ShareGPT} website for fine-tuning. The BELLE\cite{BELLE,ji2023exploringimpactinstructiondata,wen2023chathome} project's fine-tuning dataset also includes ShareGPT\cite{ShareGPT} data. User-shared data has authentic conversational context and reflects real user needs. Users tend to share high-quality, unique questions and dialogues, making user-sharing an important way to construct diverse, high-quality conversational datasets.

\subparagraph{Crowdsourcing}
Crowdsourcing has always been a common working mode for data annotation tasks. Crowdsourcing involves breaking down large-scale data tasks into smaller portions and distributing them to a large number of human workers who collaborate to process the data manually, ensuring the rapid completion of tasks. This approach has also been applied to data collection and annotation for large language model fine-tuning.\\
For instance, during the construction of the InstructGPT\cite{ouyang2022training} fine-tuning dataset, OpenAI claimed that they hired 40 annotators from a pool of candidates following specific selection criteria. This assembly of annotators formed a human data annotation team responsible for the manual generation of certain prompts and the complete generation of responses. This selection of annotators for the manual generation of fine-tuning data constitutes an effective pipeline tailored for the construction of high-quality, human-generated fine-tuning datasets.\\
Databricks-dolly-15k's\cite{databricks2023} construction process drew inspiration from the crowdsourced human data generation approach employed in InstructGPT\cite{ouyang2022training}. It involved crowdsourcing the tasks of instruction and response generation to the company's internal employees and implementing a competitive annotation mechanism where the top 20 employees with outstanding performance were eligible for rewards.\\
No Robots\cite{no_robots}, also adopts the construction design detailed in OpenAI's InstructGPT\cite{ouyang2022training} paper. As its name suggests, No Robots represents the latest progress in the human-generated fine-tuning datasets field. It creation process involved skilled human annotators.\\ 
As we can view the chat as an enhanced version of the instruction-response format. The crowdsourcing method is also utilized in the construction process of dialogue data. OpenAssistant Conversations (OASST1)\cite{köpf2023openassistant}, released by LAION AI, is a human-generated dialogue dataset with preference data developed through the crowdsourcing method, involving the participation of 13,500 volunteers globally. 

\paragraph{Model-Generated Technique} 
\subparagraph{}
We consider Self-Instruct\cite{wang-etal-2023-self-instruct} and Unnatural Instructions\cite{honovich-etal-2023-unnatural} as a turning point in the field of large model fine-tuning datasets, as they exhibit an innovative paradigm in the construction of fine-tuning datasets that save human labor during the fine-tuning phase. Self-Instruct\cite{wang-etal-2023-self-instruct}, in particular, has made a significant contribution to the field, as many subsequently released fine-tuning datasets have followed the foundational construction progress of Self-Instruct\cite{wang-etal-2023-self-instruct}. In the domain of dataset research, modifications have been made to the progress of Self-Instruct\cite{wang-etal-2023-self-instruct} to better cater to the diversity of training data requirements for models. In this section of our review, fine-tuning datasets constructed based on the Self-Instruct methodology are referred to as "Self-Instruct Style." Its pipeline consists of four components: task instruction generation, instance generation, and filtering.\\

The release of the Self-Instruct\cite{wang-etal-2023-self-instruct} open-source seed tasks and dataset has led to several subsequent fine-tuning projects adopting and expanding the Self-Instruct style. As a result, new fine-tuning datasets have been released, some of which have gained considerable influence. Stanford Alpaca\cite{taori2023alpaca} project's data generation pipeline was a modified version of the Self-Instruct generation method to build its fine-tuning datasets. Alpaca\_gpt4\_data and alpaca\_gpt4\_data\_zh, released in the Instruction Tuning with GPT-4\cite{Peng2023InstructionTW} project, are the outcomes of a modified version of the Stanford Alpaca data generation pipeline. InstructWild V1 in InstructionWild\cite{instructionwild} also used a modified version of Alpaca\cite{taori2023alpaca} data generation pipeline to generate instructions only from novel seed tasks from Twitter. BELLE\cite{BELLE}, an open-source Chinese LLM project that currently has 6k+ stars on GitHub, also includes open-source instructions. Among them is the Chinese instruction dataset BELLE/data/1.5M\cite{BELLE,ji2023exploringimpactinstructiondata,wen2023chathome}, which is generated based on the Stanford Alpaca dataset construction pipeline.\\

The model-generated method is also applied in the construction process of dialogue data. The construction pipelines of model-generated dialogue data are different, but the primary processes have points in common. Three core components are included: a) A starting point, which consists of topic seed tasks, instructions, or instruction-response pairs. b) Employed models, which are SOTA models at that time like ChatGPT. c) Designed prompts, which are important to elicit new utterances. For the construction of the dialogue data, some starting points are based on single-turn instruction data and the design of prompts is crucial for data generation.\\
ChatAlpaca.\cite{ChatAlpaca} ChatAlpaca\cite{ChatAlpaca} is constructed based on the instruction responses in the Stanford Alpaca\cite{taori2023alpaca} dataset. Through two steps involving the simulation of user utterances and model-generated responses, a multi-turn dialogue dataset is built. Unlike other multi-turn dialogue datasets, ChatAlpaca\cite{ChatAlpaca}, when utilizing model-generating capabilities, starts from existing instruction-response data, not just instructions. Through specifically designed prompts, they elicit the simulated user’s next utterance. It can be viewed as an extension of the Alpaca data in chat scenarios.\\
UltraChat.\cite{ding-etal-2023-enhancing} Its dialogue generation is accomplished by using two separate ChatGPT APIs. One API focuses on generating queries, and another is dedicated to crafting responses. The LLMs take turns working, going through many turns to make the chat data.\\
Baize.\cite{xu2023baize} Self-Chat process. Project Baize\cite{xu2023baize} defines a methodology for generating multi-turn dialogue data that employs ChatGPT to chat with itself, named Self-Chat. This naming convention has similarities with Self-Instruct. Like Self-Instruct, Self-Chat starts from seed tasks, employing existing state-of-the-art large models for corpus construction. The primary difference from Self-Instruct is that Self-Chat employs existing LLMs to engage in multi-turn dialogues, rather than merely generating instructions and single-turn instruction responses. Self-Chat can be viewed as an application of Self-Instruct in chat scenarios.\\

\paragraph{Groups of Pairing Techniques}
\subparagraph{}
Demonstration data in the generation context can be classified into three groups, based on the pairings of instruction preparation and response generation techniques. These groups are a) human-human interaction (all human labor), b) human-model interaction (dialogue with large language models), and c) model-model interaction (examples include self-instruct style or self-chat style). Regardless of the diversity in techniques or the diversity in models employed at the current era, the patterns of demonstration data generation  can be abstracted to understand and apply.\\

\paragraph{Augmentation Methods}
\subparagraph{}
Data augmentation is an important method for available public fine-tuning datasets, though a majority of demonstration datasets are currently constructed through data generation techniques. Data augmentation, as a means to enhance data diversity and improve model performance, has found widespread application in the field of machine learning. Recently generative AI has taken on an increasingly significant role in the domain of data augmentation, and this review does not delve into specific details on that matter, as that is another topic. Additionally, in our study, employing generative AI for data augmentation and the previously mentioned "Model-Generated Techniques" are two distinct concepts. We interpret the former as utilizing generative AI tools to augment data without creating entirely new data points.\\
Data augmentation methods in the construction process of available public fine-tuning datasets generally entail data sampling from existing public datasets, and when necessary, require concentrating samples from different sources, as well as applying various data augmentation techniques. Attention will be directed towards these key elements in building such datasets.\\
Firefly-train-1.1M\cite{Firefly} augmented its data by combining multiple Chinese task datasets and manually writing instruction templates, achieving a reformatting effect. Open-Platypus\cite{lee2024platypus} employs a methodology of sampling from public datasets, accompanied by content filtering, similarity exclusion, and a contamination check. MathInstruct\cite{yue2023mammothbuildingmathgeneralist} combined public datasets and prompted GPT-4 to obtain CoT (Chain of Thought) or PoT (Project of Thought) versions of existing data. Another dataset, MetaMathQA\cite{yu2023metamath}, then elaborated on their various data augmentation techniques for GSM8K\cite{cobbe2021training} and MATH\cite{hendrycksmath2021} training sets, including answer augmentation, rephrasing questions, self-verification questions and FOBAR questions.\\
We can observed that most public fine-tuning datasets obtained through data augmentation achieved a reformatted effect, which may be related to the conversational form in fine-tuning scenario. As mentioned above, large language models have also started to be used as a technical means to perform data augmentation, such as in MathInstruct\cite{yue2023mammothbuildingmathgeneralist}. \\

\subsection{Construction of Comparison Datasets}
Comparison data is more proprietary, requiring higher quality manual annotation in comparison to demonstration data. Obtaining comparison data is more challenging and costly. As a result, the availability of comparison datasets is relatively limited compared to demonstration datasets. Given the crucial role of comparison data in LLM training, let us now discuss the existing comparison datasets.\\
Constructing comparison datasets used for preference modeling in fine-tuning techniques can be abstracted into two core components.
\begin{enumerate}
\item Preparation of demonstration data
\item Preference annotation of demonstration data
\end{enumerate}

\begin{figure}[h]
    \centering
    \includegraphics[width=\linewidth]{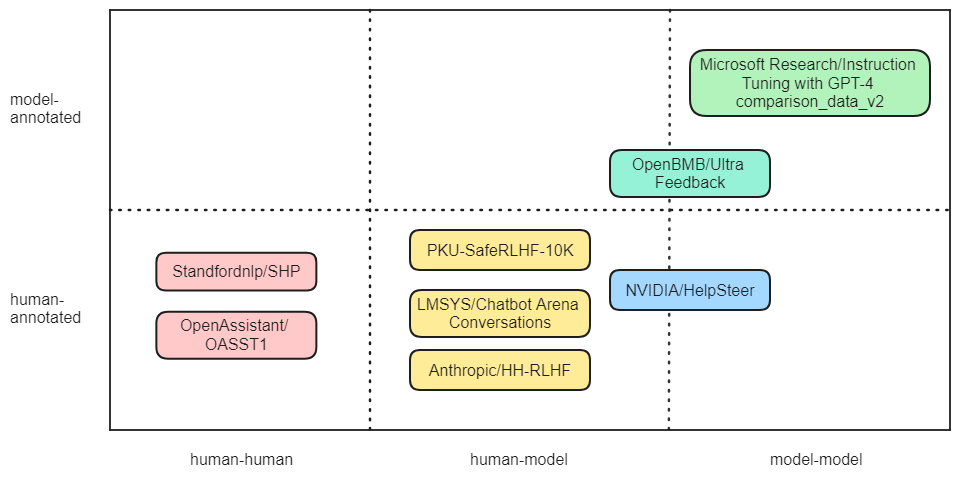}
    \caption{Construction methodologies of public comparison datasets: Two-Dimensional analysis.}
    \label{fig:6-2-dimension}
\end{figure}

The data construction in the first component does not differ significantly from the construction method of demonstration datasets, which combines both human labor and model-based work. And some of them build on open-source datasets. Distinguished from demonstration data, comparison data possess more varied types of annotation information to enhance the training of reward models. The generation of these annotations is facilitated by the preference annotation component during the construction process. In the preference annotation component, there are two distinct methods: human annotation and model-based annotation of preferences. Between them, human annotation is primarily used in existing public datasets. And a notable exception is comparison\_data in Instruction Tuning with GPT-4\cite{Peng2023InstructionTW}, which uses GPT-4 to annotate preferences. UltraFeedback\cite{cui2023ultrafeedback} Dataset serves as another instance where GPT-4 is used for preference annotation. \\
In these demonstration datasets with human annotation, the Anthropic HH-RLHF\cite{bai2022training,ganguli2022red} dataset, a foundational work in the fine-tuning field of LLMs, comprised of two parts: preference data and RedTeam data. Human preference data about helpfulness and harmlessness,  serve as comparison data for preference modeling. The human annotations of preference information in Anthropic HH-RLHF\cite{bai2022training,ganguli2022red} were obtained through sorting via an interactive interface by human annotators. The generation process of the Chatbot Arena Conversations dataset\cite{zheng2023judging}, using the interactive interface, follows a similar construction approach. Annotators utilize an online platform to pose questions and provide annotations for the corresponding models' responses. On a different note, PKU-SafeRLHF\cite{ji2023beavertails}, OpenAssistant Conversations (OASST1)\cite{köpf2023openassistant}, and HelpSteer\cite{wang2023helpsteer} share a common annotation method, which involves the utilization of crowdsourcing to obtain various preference labels. SHP\cite{pmlr-v162-ethayarajh22a} serves as an application example of an alternative method to obtain preference data. The counts of votes, likes, dislikes, and so on in forum discussion posts represent inherent preference data existing within human society, which can be collected from existing forum data for construction purposes.\\
In this review, a two-dimensional diagram\ref{fig:6-2-dimension} is employed, where the horizontal axis represents the preparation of demonstration data, and the vertical axis signifies preference annotation. This approach clearly reveals the construction methodologies of existing public comparison datasets.\\

\subsection{Construction of Generalist Datasets}
Some certain projects related to datasets in the field of fine-tuning differ from the release of specific datasets. These projects employ a methodology of dataset aggregation to produce a "generalist" dataset. OIG (Open Instruction Generalist)\cite{oig2023}, COIG (Chinese Open Instruction Generalist)\cite{zhang2023chinese}, and BELLE/data/10M\cite{BELLE,ji2023exploringimpactinstructiondata,wen2023chathome} in the BELLE project\cite{BELLE} represent this category.\\
OIG\cite{oig2023} is constructed through data augmentation, integrating information from 30 datasets, and is formatted in a conversational style. COIG dataset\cite{zhang2023chinese} follows the naming conventions and construction thoughts established by OIG\cite{oig2023}, resulting in a generalist dataset of instruction data in the Chinese domain. BELLE/data/10M dataset\cite{BELLE,ji2023exploringimpactinstructiondata,wen2023chathome}  aggregates subsets generated from the BELLE\cite{BELLE} project, using model-generated data generation techniques for their construction.

\section{Investigation Fine-tuning Datasets For MLLMs}
The pace of development in large models is remarkably rapid, marked by the emergence of iconic Multimodal Large Language Models (MLLMs) such as OpenAI's ChatGPT\cite{chatgpt} to GPT-4V\cite{yang2023dawn}, Google's release of Gemini\cite{geminiteam2024gemini}, and the recently unveiled Sora\cite{videoworldsimulators2024} by OpenAI. The enthusiasm of research institutions and individuals for exploring more multimodal large models is propelling the field towards a multimodal era. Currently, the practice of utilizing public fine-tuning datasets in the field of large-scale models is predominantly observed in the domain of Large Language Models (LLMs), with a growing practice in multimodal fine-tuning datasets as well. Up to now, public fine-tuning datasets in the multimodal domain only encompass two modalities: images and text. The construction of these image-text fine-tuning datasets can be distinguished into two main approaches.\\ Several constructive works such as InstructionBILP\cite{dai2024instructblip} and MultiInstruct\cite{xu2023multiinstruct} have also advanced the development of multimodal fine-tuning datasets. However, considering our review accords an emphasis on the concepts of public access and open-source, these specific efforts will not be elaborated on in this section.
\paragraph{Modification of Existing Computer Vision Datasets}It can be observed that the construction of most fine-tuning datasets in the multimodal domain predominantly involves a series of data modification operations based on existing public datasets for Computer vision tasks . These modification operations employ different techniques.\\ Some datasets have been developed by modifying existing datasets in the field of computer vision, utilizing captions, bounding boxes, relationships, and other textual messages representing visual information to feed language-only Large Language Models (LLMs) to create multi-modal fine-tuning datasets.LLaVA-Instruct-150K, a result of the LLaVA\cite{liu2023visual} project, supplies language-only GPT-4 with captions and bounding boxes derived from the COCO\cite{lin2014microsoft} dataset to get multimodal instruction data.GPT4RoI\cite{zhang2023gpt4roi} extends the concept of LLaVA-Instruct-150K by incorporating additional messages to address Regions of Interest. Similarly, the LAMM-Dataset\cite{yin2023lamm} employs GPT-4 to generate instructions, integrating more visual information such as relationships between objects to produce a broader variety of multi-modal fine-tuning data. The construction of the M3IT (Multi-Modal Multilingual Instruction Tuning) dataset\cite{li2023m3it} involved manual instruction writing and paraphrasing of text generated by ChatGPT.\\
Others employ multimodal models themselves to generate multimodal instructions. VIGC-InstData, originating from the VIGC\cite{wang2023vigc} framework, leverages the concept of synthetic data generation in LLMs, using Multimodal Large Language Models (MLLMs)  to create instructions based on datasets like COCO\cite{lin2014microsoft}. This generation process can be viewed as the application of Self-Instruct methodology in a multimodal scenario.
\paragraph{Generation From Scratch}In this field, both images and texts are generated through various trials. StableLLaVA\cite{li2023stablellava}, as an exemplar of this approach, employs the diffusion model to generate images directly, while utilizing the large language model ChatGPT to generate corresponding dialogue data.\\

\section{Discussion}
\paragraph{Why Public/Open-source}
Numerous public datasets have contributed to the advancement of techniques and methodologies, encompassing various dimensions of large-scale model fine-tuning, including dataset construction: a) Techniques and Methodology: Many public contributions propose innovative techniques and methodologies that span various dimensions of large-scale model fine-tuning including dataset construction, enhancing our understanding and capabilities in this area. b) Data Itself: Public datasets increasingly serve as benchmarks and sources of inspiration for proprietary fine-tuning efforts, offering a reference point for quality and diversity in data. c) Open-Source Community Development: The emergence of public fine-tuning datasets stimulates innovation and the creation of new datasets, fostering the growth and prosperity of the community.
This is precisely the focus of this review, which originates from the idea of examining public fine-tuning datasets. We aim to amplify the potential of open-source fine-tuning datasets through a systematic and general discourse, coupled with a constructive classification approach.
\paragraph{Why Construction}
The significance of construction methods in large-scale model fine-tuning data lies in their direct impact on data quality, task-specific adaptability, sample balance and diversity, as well as fine-tuning training efficiency: a) The data scale of fine-tuning datasets is often significantly smaller than that of pre-training datasets. Therefore, construction methods need to be more refined, ensuring even smaller datasets cover essential aspects of the tasks to achieve efficient learning. b) Compared to general-purpose datasets, fine-tuning datasets require a targeted representation of the specific characteristics and requirements of the task. Construction methods need to ensure a close alignment between the dataset and the task objectives, thus enhancing the effectiveness of fine-tuning. For specific domains such as healthcare and law, dataset construction needs to reflect not only the general language patterns but also domain-specific knowledge and terminologies. In such cases, the importance of construction methods becomes more prominent as they facilitate the transfer of extensive domain expertise with limited data. c) Conscious construction methods can balance datasets through sampling strategies or data augmentation, reducing biases and improving model performance and fairness in diverse environments. d) Optimizing data construction methods can reduce the required training time and resources. Selecting or generating data highly relevant to the task can improve training efficiency as models can learn more information from each sample, potentially reducing the amount of data and training cycles needed to achieve the same performance.
Construction is a priority factor in the field of fine-tuning datasets, and our review provides a reasonable induction of it to better assist researchers and organizations in deducing general principles.
\paragraph{Future Directions}
From a data modality perspective, it is anticipated that the future will witness the emergence of increasingly multifaceted fine-tuning datasets, encompassing a variety of data modalities.  The techniques and methodologies for constructing these datasets are expected to evolve alongside the rapid development of multimodal large-scale models. We anticipate a refinement and specialization in the construction of tailored datasets, catering to specific model performance characteristics such as hallucination data, security data, patch data, style-specific data, and others. The generation of these datasets will demand techniques from human contributors that are characterized by greater complexity, diversity, and higher quality, necessitating more sophisticated and enriched crowdsourcing efforts. Moreover, the utilization of model-generated synthetic data will see increased integration into the pipeline for constructing fine-tuned datasets.
\section*{Acknowledgments}
The researchers and scholars in the field whose pioneering work and valuable contributions have provided the groundwork for this review deserve our sincere appreciation. We are grateful to every community builders who has made contributions to open-sourcing the datasets, as the spirit of public and open-source support has been instrumental in completing this comprehensive review. Lastly, we would like to express our gratitude to our colleagues, OpenDataLab\cite{conghui2022opendatalab} team for their support and assistance.

\bibliographystyle{unsrt}  
\bibliography{references}

\end{document}